\documentclass{article}

\usepackage{arxiv}

\usepackage[utf8]{inputenc} 
\usepackage[T1]{fontenc}    
\usepackage{hyperref}       
\usepackage{url}            
\usepackage{booktabs}       
\usepackage{amsfonts}       
\usepackage{nicefrac}       
\usepackage{microtype}      
\usepackage{lipsum}
\usepackage[bottom]{footmisc} 
\usepackage{graphicx}
\graphicspath{ {./images/} }
\usepackage{authblk}  
\title{Do Deepfake Detectors Work in Reality?}
\author[1]{Simiao Ren\textsuperscript{\dag}}
\author[2]{Hengwei Xu\textsuperscript{*}}
\author[4]{Tsang (Dennis) Ng\textsuperscript{*}}
\author[4]{Kidus Zewde\textsuperscript{*}}
\author[4]{Shengkai Jiang\textsuperscript{*}}
\author[4]{Ramini Desai\textsuperscript{*}}
\author[4]{Disha Patil\textsuperscript{*}}
\author[4]{Ning-Yau Cheng\textsuperscript{*}}
\author[3]{Yining Zhou\textsuperscript{*}}
\author[4]{Ragavi Muthukrishnan\textsuperscript{*}}

\affil[1]{Duke University, \texttt{simiao.ren@duke.edu}}
\affil[2]{Georgia Tech, \texttt{hxu457@gatech.edu}}
\affil[3]{Texas A\&M, \texttt{xwyzyn135@gmail.com}}
\affil[4]{Scam.ai, \texttt{\{dennis.ng, kidus.zewde, shengkai.jiang, ramani.desai, disha.patil, rachel.cheng, ragavi.muthukrishnan\}@scam.ai}}

\begin{document}
\maketitle
{\renewcommand{\thefootnote}{}\footnote{Authorship of this paper does not imply employment by or affiliation with Scam.AI (Reality Inc.); some authors are independent researchers who collaborated on this work.}\addtocounter{footnote}{-1}}
\begin{abstract}

 Deepfakes, particularly those involving faceswap-based manipulations, have sparked significant societal concern due to their increasing realism and potential for misuse. Despite rapid advancements in generative models, detection methods have not kept pace, creating a critical gap in defense strategies. This disparity is further amplified by the disconnect between academic research and real-world applications, which often prioritize different objectives and evaluation criteria. In this study, we take a pivotal step toward bridging this gap by presenting a novel observation: the post-processing step of super-resolution, commonly employed in real-world scenarios, substantially undermines the effectiveness of existing deepfake detection methods. To substantiate this claim, we introduce and publish the first real-world faceswap dataset, collected from popular online faceswap platforms. We then qualitatively evaluate the performance of state-of-the-art deepfake detectors on real-world deepfakes, revealing that their accuracy approaches the level of random guessing. Furthermore, we quantitatively demonstrate the significant performance degradation caused by common post-processing techniques. By addressing this overlooked challenge, our study underscores a critical avenue for enhancing the robustness and practical applicability of deepfake detection methods in real-world settings.

\end{abstract}

\section{INTRODUCTION}

\renewcommand{\thefootnote}{\fnsymbol{footnote}}

\footnotetext[2]{\dag\ Correspondence author.}
\footnotetext[1]{* Equal contributions, order randomly generated.}

The rise of artificial intelligence has benefited various fields from material design to energy\cite{ren2020benchmarking, ren2022automated, ren2024segment, mandal2020acoustic}. However, among those achievements, face-swap technology has emerged as a double-edged sword, showcasing remarkable advancements in artificial intelligence while simultaneously posing significant ethical and societal challenges \cite{westerlund2019emergence}. By seamlessly superimposing one individual’s face onto another’s body in videos, this technology blurs the line between reality and fabrication, undermining the trust that forms the foundation of modern society. The malicious use of deepfake face-swapping to create deceptive media—ranging from non-consensual explicit content \cite{karasavva2021real} to fraudulent political campaigns \cite{pantserev2020malicious}—has eroded public confidence in the authenticity of digital content. As these forgeries become increasingly indistinguishable from genuine media, they foster skepticism and paranoia, threatening interpersonal relationships, organizational credibility, and democratic processes.

Despite the recognized dangers of deepfake face-swap technology, current detection mechanisms face significant limitations, particularly in real-world applications. While many detection algorithms achieve high accuracy in controlled laboratory conditions \cite{yan2023deepfakebench}, their performance often degrades when applied to real-world data. Beautification filters and post-processing techniques, commonly applied to media in practical scenarios, exacerbate this challenge by obscuring the subtle artifacts that detection systems rely on. This gap between theoretical robustness and practical reliability underscores the urgent need to better understand how real-world deepfakes differ from academic datasets and to quantitatively analyze the reliability of detection methods in real-world settings. Without such advancements, the growing prevalence of undetectable deepfakes will continue to erode trust in digital content, amplifying their societal impact.

In this work, we address this gap by conducting rigorous experiments that demonstrate how current deepfake detectors fail in real-world settings and exploring the reasons behind this failure. Our contributions are fourfold:

\begin{itemize}
    \item \textbf{Creating and publishing a real-world faceswap dataset}:We hand-collected over 800 faceswap images from top-ranked online faceswap tools identified via search engines. This dataset represents the distribution of deepfakes most likely encountered by the general public.
    \item \textbf{Benchmark current state-of-the-art faceswap detection models on real-world faceswap dataset}:We evaluated several state-of-the-art deepfake detectors on this dataset, revealing their vulnerabilities when faced with real-world data, despite achieving high accuracy in academic settings.
    \item \textbf{Crucial observation that post-processing step of super-resolution contributes significantly to fooling deepfake detectors}: Through reverse engineering, we discovered that the post-processing step of super-resolution significantly contributes to the failure of deepfake detectors by introducing a distribution shift.
    \item \textbf{Quantitative analysis on performance degradation of deepfake detectors facing super-solution and filters}: We conducted a detailed quantitative analysis to benchmark the performance degradation of deepfake detectors when subjected to various super-resolution techniques and beautification filters.

\end{itemize}

\section{Related work}

The rapid development of public deepfake datasets, such as FaceForensics++ (FF++) \cite{rossler2019faceforensics++}, CelebDF \cite{li2020celeb}, DFD \cite{deepfake-detection-challenge}, and DFDC \cite{DFDC2020}, has provided diverse test grounds for evaluating deepfake detection methods. Recent benchmark studies \cite{yan2023deepfakebench, li2023continual} have pointed out naive detectors directly train models on labeled datasets using architectures like Xception \cite{chollet2017xception} and EfficientNet \cite{tan2019efficientnet, shiohara2022detecting} have comparable performances to spatial detectors.

Model-based image upscaling harms spatial detectors that rely on blending artifacts \cite{li2020face, nguyen2024laa, shiohara2022detecting, li2018exposing}. Unlike previous studies, we focus on the post-processing step where super-resolution models, such as CodeFormer \cite{zhou2022towards}, enhance deepfake outputs. While some detection methods \cite{liu2021spatial, tan2024rethinking} target upscaling in encoder-decoder structures, our work highlights the challenge posed by post-processing upscaling, which introduces distribution shifts that significantly degrade detection performance.

Previous studies \cite{libourel2024case} have examined the adverse effects of post-processing on deepfake detection, particularly focusing on beautification filters commonly used on social media. However, our work diverges by addressing a more pervasive real-world phenomenon: model-based upscaling.

\addtolength{\textheight}{-3cm}   

\section{Methodology}

\subsection{Real-world faceswap (RWFS) dataset creation}

\begin{figure}[h!]
    \centering
    \includegraphics[width=0.8\textwidth]{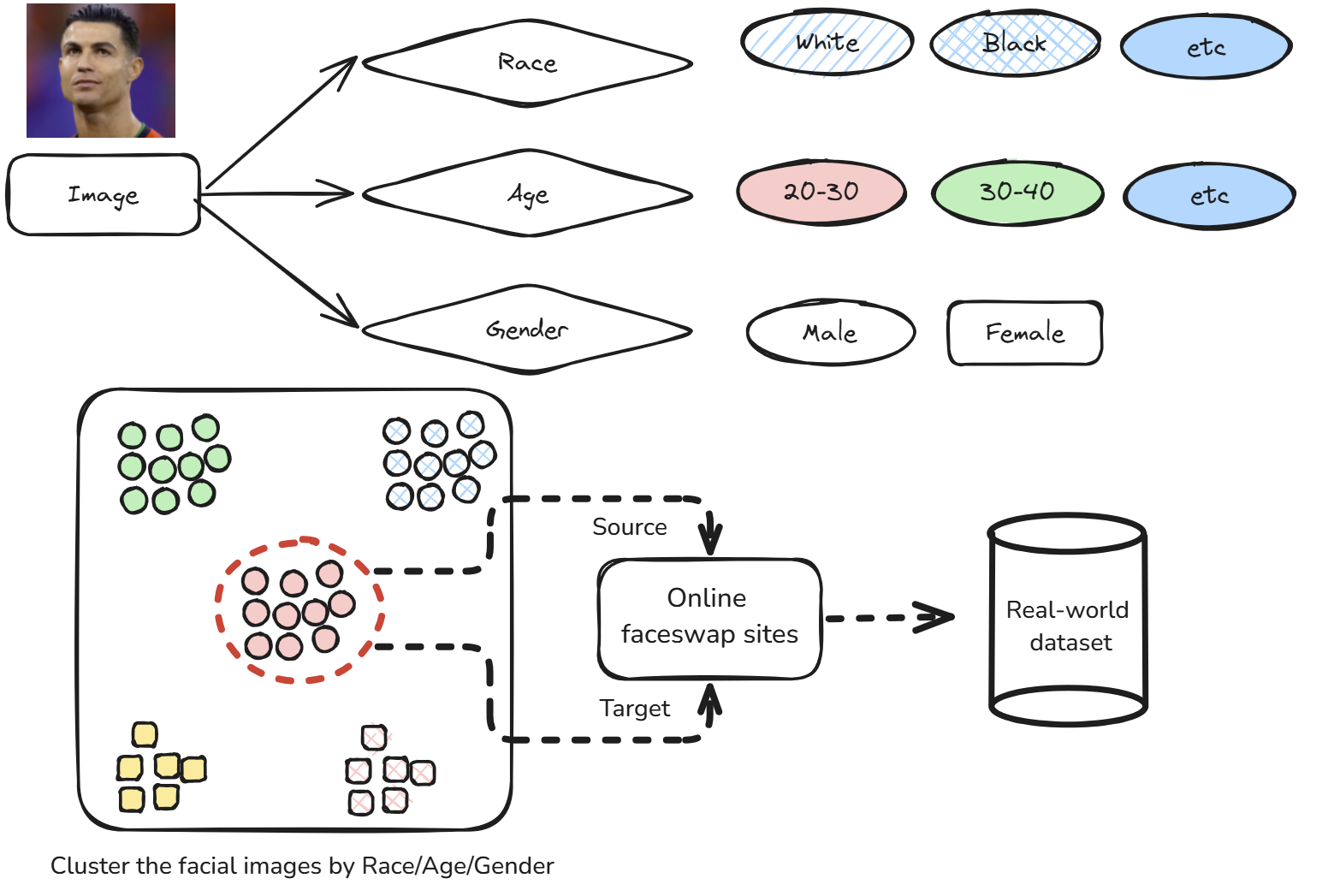} 
    \caption{Real-world faceswap dataset generation process with race-age-gender matching}
    \label{fig:dataset_generation}
\end{figure}

During our initial investigations, we observed that the quality of real-world deepfakes found on social media and in the news significantly surpasses that of those in existing academic benchmark datasets. This observation motivated us to create a new deepfake detection dataset that better reflects real-world settings. To achieve this, we introduced two key distinctions from traditional academic datasets in our creation process:

\begin{itemize}
    \item \textbf{Utilizing the top online deepfake generators for real world application}:While there is an abundance of face-swap repositories on GitHub and methodologies in academic literature, we deliberately avoided these sources for dataset creation. These methods are often more "academic" in nature and do not represent the tools used by the general public. In real-world scenarios, a bad actor seeking to generate deepfake imagery is more likely to rely on freely available online resources. To replicate this behavior, we identified and used the top eight websites returned by searching "faceswap, free" that allow users to generate faceswap images at no cost.
  \item \textbf{Matching race-gender-age pairs of our source and target faces}: A common limitation of academic benchmark datasets is the low fidelity resulting from significant mismatches between the source and target facial features. In real-world scenarios, a bad actor is unlikely to swap between two drastically different faces. To address this, we implemented a "Race-Gender-Age" matching mechanism. Using race, age, and gender prediction models, we ensured that face swaps were performed only between individuals with matching characteristics. This approach enhances the realism of our dataset and better aligns with real-world deepfake practices, as illustrated in Figure \ref{fig:dataset_generation}.
\end{itemize}

We also evaluate contemporary deepfake detectors on our newly collected real-world dataset to highlight the gap between academic research and real-world application scenarios. Specifically, we selected the naive EfficientNet-B4 detector from a recent benchmark study \cite{yan2023deepfakebench}, which demonstrated strong overall performance across various datasets. Additionally, we included the self-blended images approach \cite{shiohara2022detecting}, known for its exceptional generalization to unknown scenarios, to assess its robustness in real-world settings.

\subsection{Reverse engineer post processing step through self-swap}
After collecting our real-world deepfake dataset, we observed intriguing artifacts that were absent in the input imagery. To further investigate how the processing in real-world datasets differs from that in academic datasets, we devised a "self-swap" test. This test involved reverse-engineering the post-processing pipeline of online deepfake generators by swapping a face with itself. Under the assumption that swapping a face with itself should produce identical imagery, we discovered that this assumption was violated across all tested online deepfake sites. This finding indicates the presence of additional post-processing transformations.

Through careful observation and unsuccessful attempts to replicate these transformations using facial smoothing and beautification filters, we identified that the applied transformations are model-based super-resolution techniques. These techniques introduce subtle yet significant changes to the imagery, distinguishing real-world deepfakes from their academic counterparts.


\subsection{Quantitatively evaluation of super-resolution impact}
To test the hypothesis that post-processing steps involving super-resolution contribute to the degradation of deepfake detection accuracy, we conducted a quantitative evaluation comparing the performance of state-of-the-art deepfake detectors on standard benchmark imagery versus super-resolved imagery. For this analysis, we selected multiple state-of-the-art deepfake detectors and super-resolution methods.

We chose GFPGAN \cite{wang2021gfpgan} and CodeFormer \cite{zhou2022codeformer} for our experiments due to their popularity and demonstrated robustness. GFPGAN has garnered over 10 million downloads on GitHub, reflecting its widespread adoption, while CodeFormer is the latest state-of-the-art method, achieving superior visual quality in several benchmarks. Both methods provide a strong basis for evaluating the impact of super-resolution on deepfake detection performance.

\begin{table}[h!]
\centering
\caption{Number of images generated from each online faceswap resource in the Real-World Faceswap Dataset.}
\label{tab:dataset_sources}
\begin{tabular}{ccc}
\toprule
\textbf{Source Website} & \textbf{URL} & \textbf{Number of Images} \\ \midrule
Pixlr                  & \href{https://pixlr.com/face-swap/}{pixlr.com/face-swap} & 81 \\ 
Magic Hour             & \href{https://magichour.ai/products/face-swap}{magichour.ai/products/face-swap} & 104 \\ 
Remaker                & \href{https://remaker.ai/face-swap-free/}{remaker.ai/face-swap-free} & 92 \\ 
AI FaceSwap IO         & \href{https://aifaceswap.io/}{aifaceswap.io} & 71 \\ 
Ismarta                & \href{https://www.ismartta.com/}{ismartta.com} & 93 \\ 
Pica                   & \href{https://www.pica-ai.com/}{pica-ai.com} & 84 \\ 
Widwud                 & \href{https://www.vidwud.com/free-face-swap}{vidwud.com/free-face-swap} & 95 \\ 
Faceswapper            & \href{https://faceswapper.ai/}{faceswapper.ai} & 227 \\ \midrule
\textbf{Total}         & --- & \textbf{847} \\ 
\bottomrule
\end{tabular}
\end{table}

\section{Results}

\subsection{RWFS dataset Created}

We created 848 real-world faceswap images using eight online faceswap websites, with the number of images from each site shown in Table \ref{tab:dataset_sources}. For the real images in the RWFS dataset, we randomly selected 900 from the Celeb dataset (the same source as the fake images).

We also evaluated popular deepfake detection algorithms on this dataset. While these detectors perform well on benchmark datasets like FF++ and CelebDF, they fail significantly on our RWFS dataset. We tested two pretrained networks that have shown strong performance in prior studies to demonstrate this.

\begin{figure}[h!]
    \centering
    \includegraphics[width=0.8\textwidth]{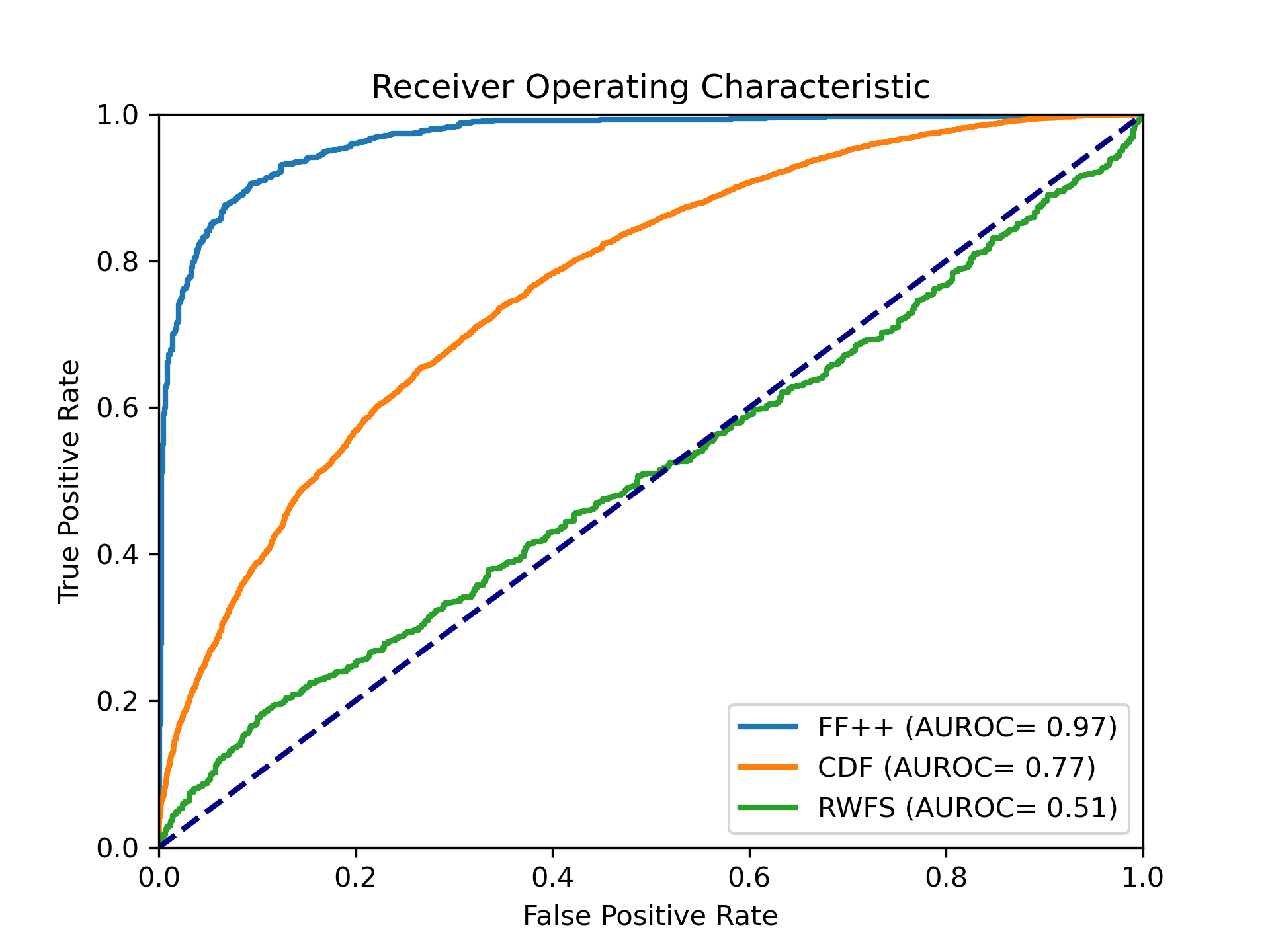} 
    \caption{Efficient-b4 naive detector pretrained on FF, weights taken from \cite{yan2023deepfakebench}.}
    \label{fig:perf_sbi}
\end{figure}

\begin{figure}[h!]
    \centering
    \includegraphics[width=0.8\textwidth]{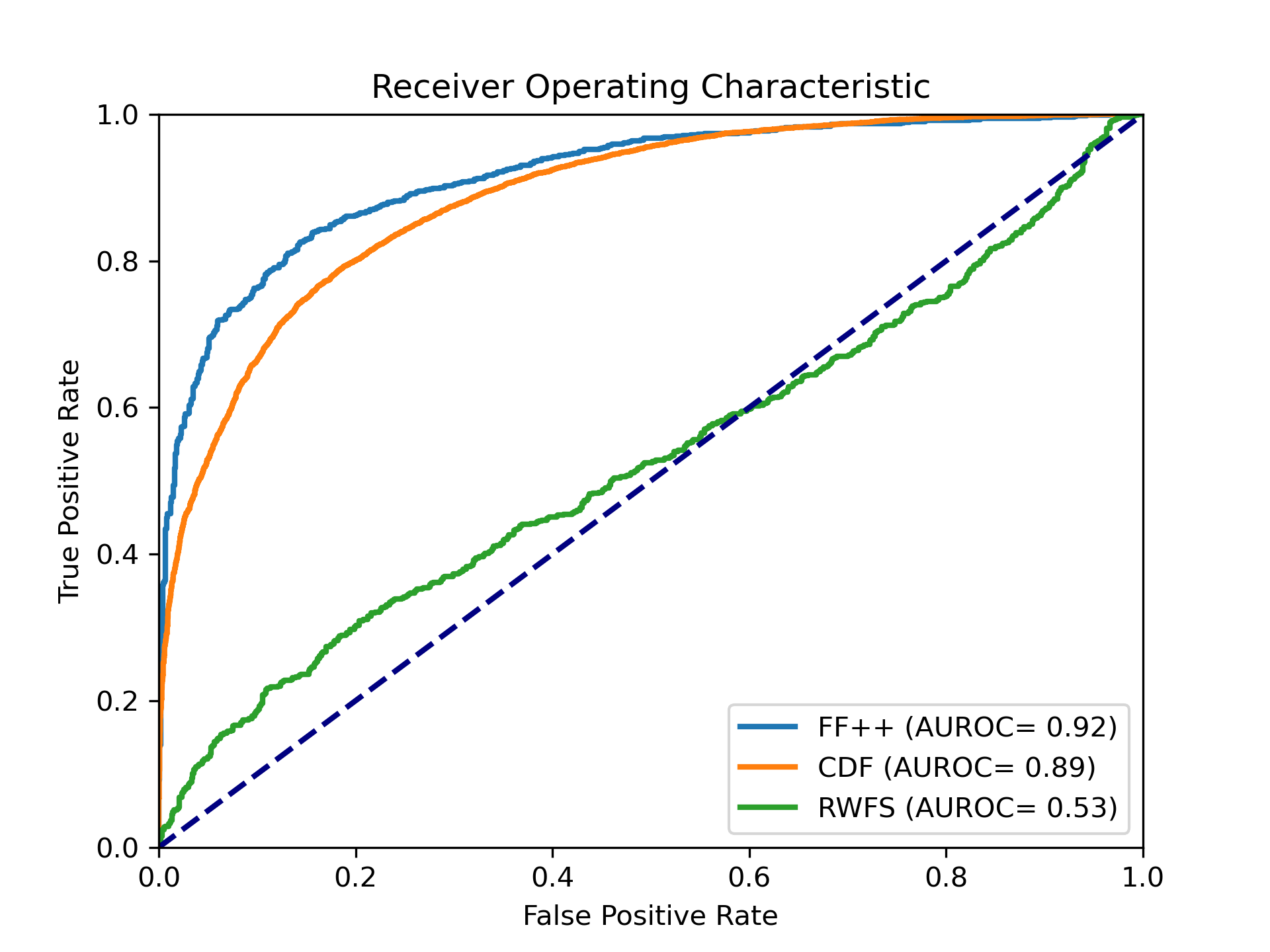} 
    \caption{Self-blended imagery detector pretrained on FF, weights taken from \cite{chen2022self}}
    \label{fig:perf_efficient_b4}
\end{figure}

To ensure a fair comparison, we reproduced their scores in our test environment and reported both the reproduced performance and the originally published performance. Figures \ref{fig:perf_efficient_b4} and \ref{fig:perf_sbi} illustrate the stark contrast: academically successful networks fail to generalize to real-world image distributions. Specifically, pretrained models from FF++ achieve AUROC scores exceeding 0.92 on balanced academic test sets but perform as poorly as 0.53 (close to random guessing) on our real-world dataset. This discrepancy underscores the significant gap between what academia is optimizing for and the requirements of real-world applications.

\subsection{Self-swap results}
\begin{figure}[h!]
    \centering
    \includegraphics[width=0.45\textwidth]{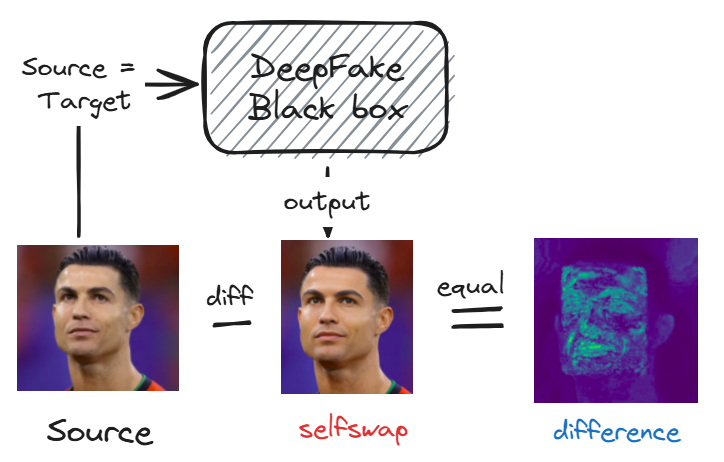} 
    \caption{Visualizing the self-swap result by plotting the difference.}
    \label{fig:self_swap_result}
\end{figure}

We first compared the self-swapped images with the original face images by plotting their differences, as shown in Figure \ref{fig:self_swap_result}. Upon closer inspection, we observed distinct changes in the self-swapped images. For instance, in the original image, the double eyelids were barely visible, and the eyelashes were clearly defined. However, in the self-swapped images, the double eyelids became significantly more pronounced, and the reflections in the eyes changed noticeably.

These changes suggest that the transformations go beyond simple beautification filters (such as those investigated by Libourel et al. \cite{libourel2024case}), which primarily smooth the skin and adjust color scales. Instead, they indicate the application of advanced deep processing techniques that introduce new details into the imagery. Further comparisons with popular model-based image super-resolution techniques confirmed that these artifacts are characteristic of such post-processing steps.

\subsection{Super-resolution}

\begin{figure}[h!]
    \centering
    \includegraphics[width=0.8\textwidth]{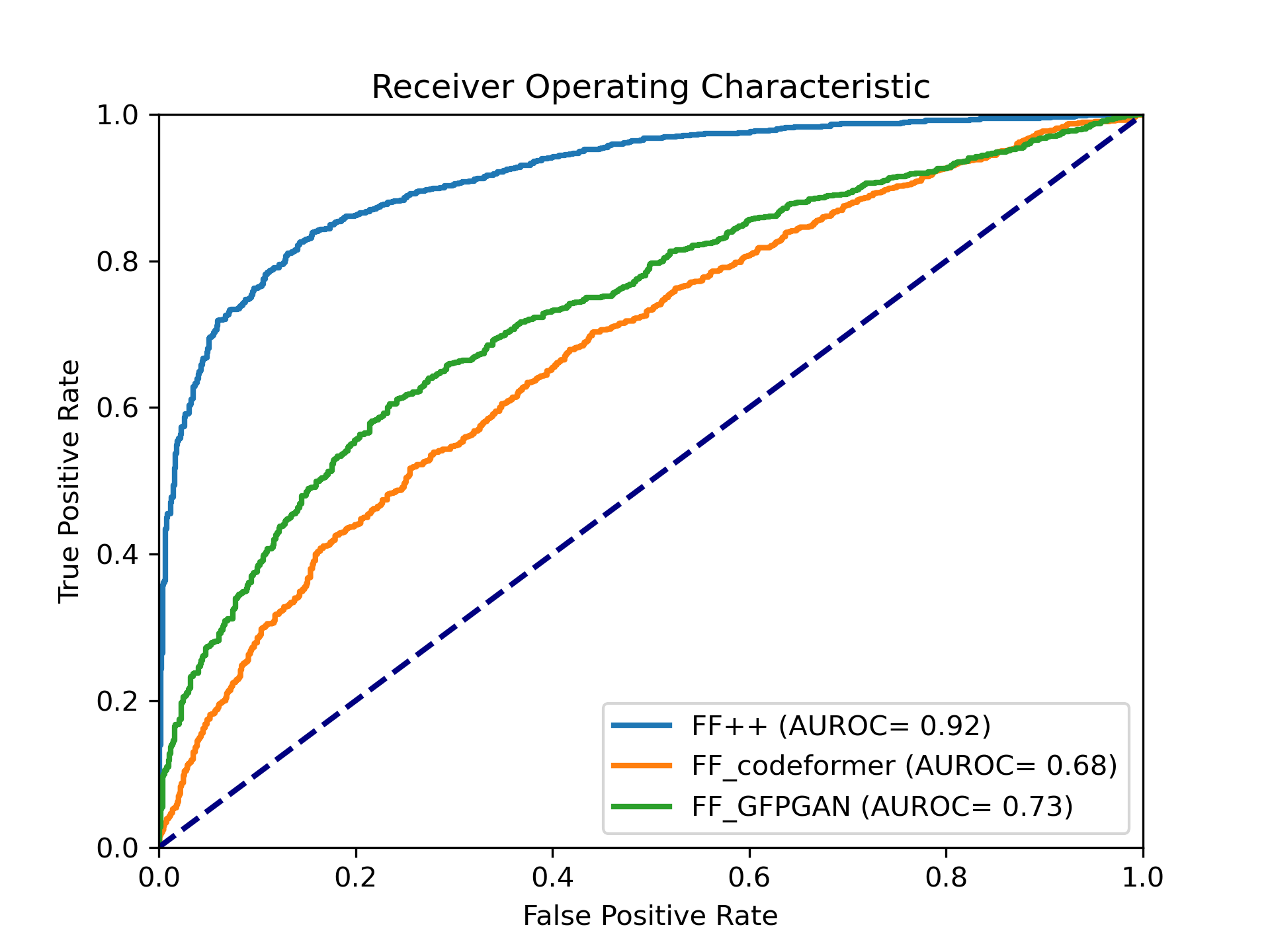} 
    \caption{Self-blended images model performance degradation caused by super-resolution on FF++}
    \label{fig:sbi_super_res}
\end{figure}

\begin{figure}[h!]
    \centering
    \includegraphics[width=0.8\textwidth]{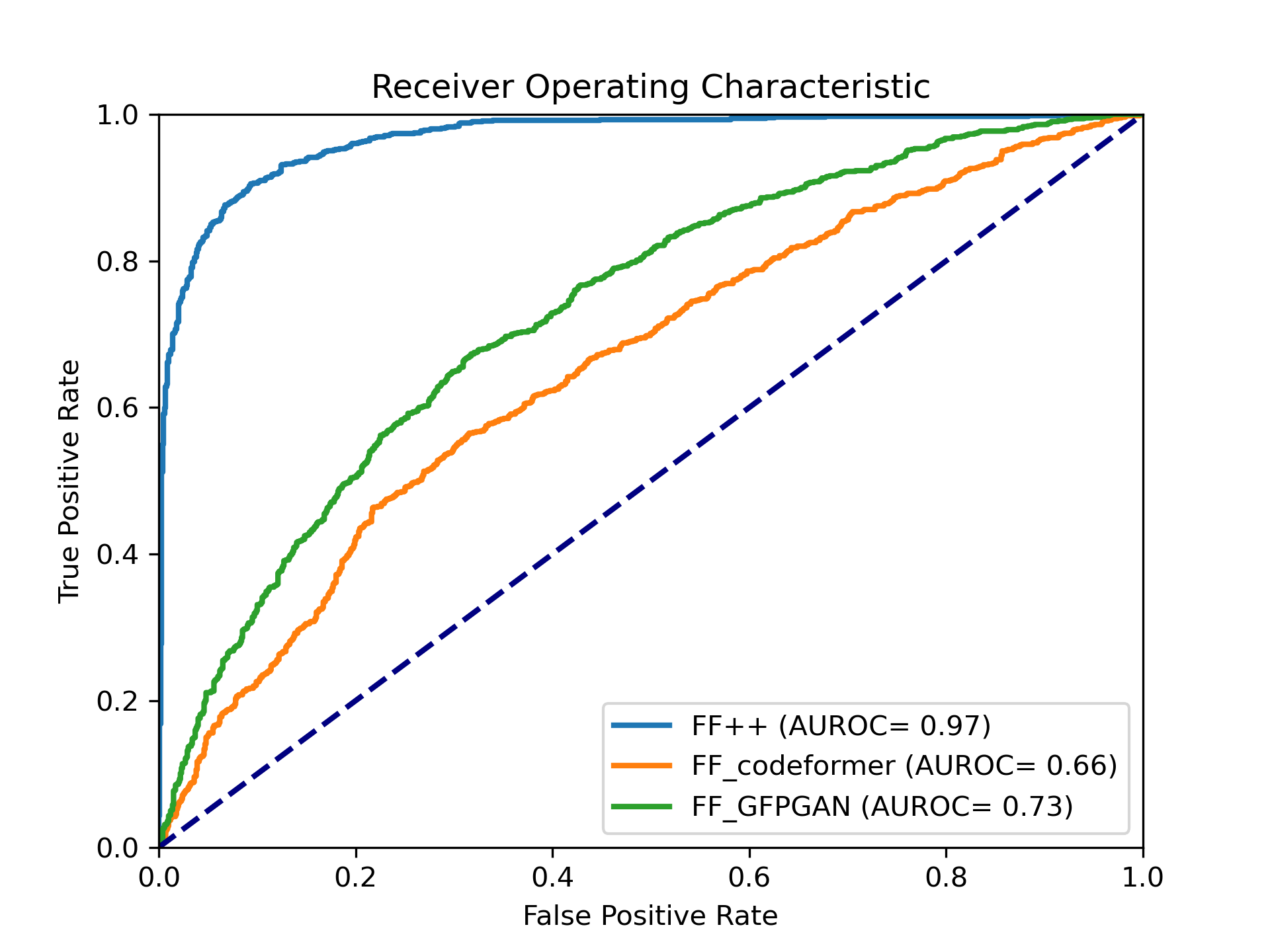} 
    \caption{naive efficient-b4 model performance degradation caused by super-resolution on FF++}
    \label{fig:efficient_super_res}
\end{figure}

With super-resolution identified as a major factor contributing to the degradation of deepfake detection performance, we applied super-resolution algorithms directly to the FF++ dataset and evaluated the resulting performance degradation. The results are presented in Figure \ref{fig:sbi_super_res} and Figure \ref{fig:efficient_super_res}. From these figures, it is evident that the performance of deepfake detectors is significantly reduced by the application of super-resolution algorithms. Specifically, the AUROC scores dropped from over 0.9 to approximately 0.7 for both the self-blended images model and the Efficient-B4 naive model.

Moreover, the extent of performance degradation varies depending on the specific super-resolution algorithm used. This discrepancy can be attributed to the fact that deepfake detector models are typically trained to identify specific artifacts, such as blending inconsistencies at face edges. When bad actors introduce an additional processing step, such as using another neural network-based super-resolution algorithm like GFPGAN or CodeFormer, the original model artifacts are significantly diminished or replaced with new traces introduced by the super-resolution model. This shift in artifact patterns confuses the deepfake detectors, leading to reduced performance.

It is important to note, however, that the performance degradation caused by super-resolution does not yet fully account for the challenges posed by our RWFS dataset. Upon closer examination of the super-resolution outputs, we observed that many still exhibit artifacts that are easily detectable by humans. This is likely because the original FF++ dataset contains highly corrupted and low-quality images (from a human perception perspective) compared to the higher-quality real-world images in our RWFS dataset.

\section{Conclusions}
Deepfake face-swap technology poses significant social risks by undermining trust in digital media. While detection algorithms perform well in controlled settings, their effectiveness diminishes in real-world scenarios due to post-processing techniques like super-resolution.

In this work, we introduced the Real-World Faceswap Dataset, the first to reflect public-facing deepfakes, and benchmarked state-of-the-art detectors on it. Our results highlight vulnerabilities in current detection methods, particularly the impact of super-resolution and post-processing on performance.

By providing this dataset and insights, we aim to bridge the gap between academic research and practical applications, advancing deepfake detection to mitigate their societal impact.

\section*{Data and Code Availability}
The datasets and code supporting this work are released by Scam.AI and are available from
the Scam.AI Research portal, \url{https://scam.ai/research}~\cite{scamai_research}. Document-
and PDF-forgery resources are additionally maintained at Trustable, \url{https://trustable.ai}~\cite{trustable_ai}.
These are the canonical, maintained locations for the resources associated with this paper and
supersede any prior ad hoc distribution.

\bibliographystyle{plain}
\bibliography{reference}
\end{document}